\documentclass[sigconf]{acmart}
\usepackage[utf8]{inputenc} 
\usepackage[T1]{fontenc}    
\usepackage{textcomp}       
\DeclareUnicodeCharacter{266B}{\textmusicalnote}
\usepackage[table]{xcolor}
\usepackage{hyperref}       
\usepackage{url}            
\usepackage{booktabs}       
\usepackage{tabularx}       
\usepackage{amsfonts}       
\usepackage{nicefrac}       
\usepackage{microtype}      
\usepackage{xcolor}         

\usepackage{amsmath}

\usepackage{amssymb}
\usepackage{multirow}
\usepackage{graphicx}
\usepackage{pifont}
\usepackage{bm}         
\usepackage{mathtools}  
\usepackage{bbm}        
\usepackage{algorithm}
\usepackage{algpseudocode}
\usepackage{float} 
\usepackage{graphicx}
\usepackage{subcaption}

\AtBeginDocument{%
  }

\setcopyright{acmlicensed}
\copyrightyear{2018}
\acmYear{2018}
\acmDOI{XXXXXXX.XXXXXXX}

\acmConference[Conference acronym 'XX]{Make sure to enter the correct
conference title from your rights confirmation email}{June 03--05,
2018}{Woodstock, NY}
\acmISBN{978-1-4503-XXXX-X/2018/06}

\begin{document}

\title{\textsc{TaSR-RAG}: Taxonomy-guided Structured Reasoning for
Retrieval-Augmented Generation}


\author{
Jiashuo Sun, Yixuan Xie, Jimeng Shi, Shaowen Wang, Jiawei Han
}
\affiliation{%
\institution{University of Illinois Urbana-Champaign}
  \country{}
}
\email{
{jiashuo5, hanj}@illinois.edu
}


\renewcommand{\shortauthors}{Trovato et al.}

\begin{abstract}


  Retrieval-Augmented Generation (RAG) helps large language models (LLMs) answer knowledge-intensive and time-sensitive questions by conditioning generation on external evidence. However, most RAG systems still retrieve unstructured chunks and rely on one-shot generation, which often yields redundant context, low information density, and brittle multi-hop reasoning. While structured RAG pipelines can improve grounding, they typically require costly and error-prone graph construction or impose rigid entity-centric structures that do not align with the query's reasoning chain.
  We propose \textsc{TaSR-RAG}, a taxonomy-guided structured reasoning framework for evidence selection. We represent both queries and documents as relational triples, and constrain entity semantics with a lightweight two-level taxonomy to balance generalization and precision. Given a complex question, \textsc{TaSR-RAG} decomposes it into an ordered sequence of triple sub-queries with explicit latent variables, then performs step-wise evidence selection via hybrid triple matching that combines semantic similarity over raw triples with structural consistency over typed triples.
  By maintaining an explicit entity binding table across steps, \textsc{TaSR-RAG} resolves intermediate variables and reduces entity conflation without explicit graph construction or exhaustive search. Experiments on multiple multi-hop question answering benchmarks show that \textsc{TaSR-RAG} consistently outperforms strong RAG and structured-RAG baselines by up to 14\%, while producing clearer evidence attribution and more faithful reasoning traces.
\end{abstract}




\received{20 February 2007}
\received[revised]{12 March 2009}
\received[accepted]{5 June 2009}

\maketitle

\section{Introduction}

Large language models (LLMs) have demonstrated strong performance on knowledge-intensive tasks, yet their parametric knowledge alone is often insufficient for answering factual or time-sensitive questions \cite{kilt, rag_survey}. Retrieval-Augmented Generation (RAG) mitigates this limitation by retrieving external documents and conditioning generation on the retrieved evidence \cite{ram-etal-2023-context, rag-task}. In principle, this paradigm can provide grounded answers with citations and enable models to access up-to-date information beyond training data. However, most RAG systems still operate on unstructured text chunks and rely on one-shot prompting, which frequently leads to redundant context, low information density, and brittle multi-hop reasoning \cite{rankrag, replug}. These weaknesses are amplified for multi-hop queries, where the evidence is scattered across documents and intermediate conclusions must be carried forward to answer the final question, often under tight context-window budgets.

Recent structured RAG methods incorporate knowledge graphs or extracted triples to improve factual grounding \cite{graphrag, hyper-cube-rag, hypergraphrag, hipporag, structrag, structure-r1}. Nevertheless, graph-based pipelines often require expensive construction and may introduce hallucinated nodes or misaligned structure, while entity-centric triple extraction can be overly sparse and poorly aligned with the query's reasoning chain. More broadly, existing structured pipelines often lack a principled way to jointly leverage semantic relevance (what the query asks for) and structural compatibility (whether the retrieved facts satisfy the required entity types and relations) when selecting evidence across hops.

\begin{figure}[t]
  \centering
  \includegraphics[width=0.49\textwidth]{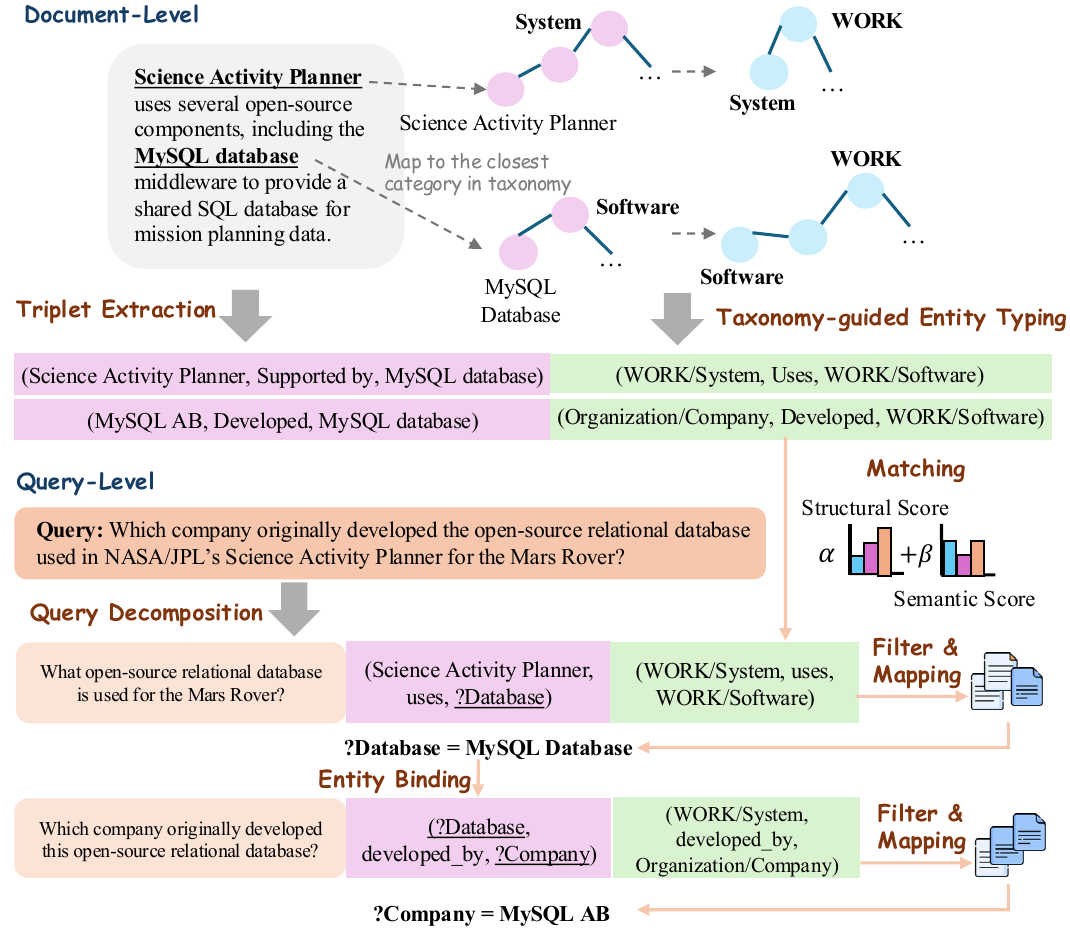}
  \caption{Overview of \textsc{TaSR-RAG}. We illustrate Document-level triple extraction and taxonomy-guided entity typing, query decomposition with latent variables, hybrid triple matching for document reranking, and sequential entity binding.}
  \label{fig:intro}
\end{figure}

We argue that the core challenge is multi-hop evidence selection: selecting and organizing evidence in the order needed to resolve intermediate unknowns.
Considering a compositional question in Figure \ref{fig:intro}: ``Which database does \textit{Science Activity Planner} use, and which company developed it?'' a standard one-shot RAG pipeline may retrieve passages that mention multiple databases and vendors, then generate a plausible but weakly supported answer due to entity conflation.
Ideally, the system should first resolve the intermediate unknown (the database) from direct evidence (e.g., ``\textit{Science Activity Planner uses MySQL database}''), and then use that resolved entity to guide retrieval for the next hop (who developed \textit{MySQL}).
To make this step-wise evidence selection reliable, we should match facts by jointly considering semantic similarity over raw triples and structural consistency over typed triples (e.g., \textsc{WORK/System} $\rightarrow$ \textsc{WORK/Software} $\rightarrow$ \textsc{Organization/Company}).

Based on this motivation, we propose \textbf{TaSR-RAG}, a taxonomy-guided sequential reasoning framework that represents both documents and queries as relational triples and augments them with lightweight two-level taxonomy types.
\textsc{TaSR-RAG} decomposes the query into an ordered sequence of triple-form sub-queries with latent variables (e.g., $s_1=(\text{Science Activity Planner},\allowbreak\,\texttt{uses},\allowbreak\,\text{?Database})$, $s_2=(\text{?Database},\allowbreak\,\texttt{developed\_by},\allowbreak\,\text{?Company})$) and performs step-wise context selection with explicit bindings.
At each step, we rerank candidate documents with a hybrid (semantic + structural) triple matching function that enforces consistency over both raw triples and taxonomy-typed triples, and then update bindings; this enables later retrieval and generation to be conditioned on earlier resolutions and improves faithfulness over one-shot RAG. This design yields an efficient and interpretable RAG system that produces explicit reasoning traces (sub-queries, matched triples, and bindings) without explicit graph construction. Because each step produces a small set of type-consistent candidate facts, later LLM calls are less likely to be distracted by irrelevant context and can be audited using the intermediate bindings. Empirically, \textsc{TaSR-RAG} consistently outperforms strong baselines across seven QA benchmarks. On \texttt{Qwen2.5-72B-Instruct}, we improve the average EM from 29.7 (standard RAG) to 42.5. On \texttt{Qwen2.5-7B-Instruct}, we improve the average EM from 21.1 (standard RAG) to 37.0, and achieve up to a 103\% relative gain on Musique compared to the strongest baseline.

Our contributions are threefold: (1) we introduce a taxonomy-guided typed-triple representation for both documents and queries; (2) we propose a hybrid matching function that combines semantic similarity over raw triples with structural consistency over typed triples; and (3) we develop a sequential context selection procedure with explicit latent-variable binding for interpretable multi-hop reasoning. Notably, \textsc{TaSR-RAG} is training-free and can be integrated with existing dense retrievers and LLMs as a modular reranking-and-reasoning layer.

\section{Related Work}

\subsection{Retrieval-Augmented Generation}

RAG enhances LLM performance by incorporating external evidence, improving factual grounding and reducing hallucination in knowledge-intensive tasks \cite{rag_survey, Guu2020RAG}. Beyond chunk-based retrieval, work on retrieval-augmented next-token prediction \cite{rag-task, kilt} and end-to-end training frameworks \cite{replug, dynamicrag, grace, sun2026rethinkingrerankerboundaryawareevidence} aims to tighten the integration between retriever and generator. Self-reflective RAG systems such as \citet{selfrag} employ special control tokens and a critic model to adaptively trigger retrieval and improve reasoning stability.

While these efforts address retrieval quality or retriever--generator integration, our work targets a different bottleneck: multi-hop context selection. Instead of retrieving unstructured passages and delegating all reasoning to the LLM, \textsc{TaSR-RAG} represents both documents and queries as relational triples and augments them with taxonomy-guided entity types. This enables a hybrid matching function that combines semantic similarity over raw triples with structural consistency over typed triples, and supports sequential reasoning via explicit latent-variable binding.

\subsection{Structure-aware Retrieval and Reasoning}

A complementary line of work introduces structural supervision into retrieval through knowledge graphs, LLM-generated topic graphs, or symbolic abstractions \cite{tog}. GraphRAG \cite{graphrag}, HippoRAG \cite{hipporag}, and HyperGraphRAG \cite{hypergraphrag} construct large hierarchical graphs from LLM-generated summaries, supporting multi-hop reasoning but suffering from hallucinated nodes, misaligned structure, and prohibitive construction cost. StructRAG \cite{structrag} and Structure-R1 \cite{structure-r1} and related triple-based systems extract entity-level triples to reduce noise, yet these triples remain sparse, overly fine-grained, and lack structural alignment with the query, limiting multi-hop generalization. Ontology-based information extraction has also been explored in traditional IR and semantic parsing, but prior methods typically assume fixed schemas and do not integrate typed reasoning patterns from the query itself.

Our approach differs by avoiding explicit graph construction while still exploiting structure. \textsc{TaSR-RAG} converts retrieved documents into triples and their corresponding taxonomy-typed triples, decomposes the query into typed sub-queries with latent variables, and reranks documents at each step using hybrid (semantic + structural) triple matching. By iteratively answering sub-queries and binding intermediate entities, our method aligns the evidence selection process with the query's reasoning chain, yielding interpretable reasoning traces without relying on hallucinated graph nodes or expensive graph-building pipelines.

\section{Method}

\subsection{Design Motivation}

The key insight behind TaSR-RAG is that complex multi-hop queries require structured reasoning over retrieved documents, yet existing RAG systems typically rely on flat, query-document similarity that ignores the relational structure of information needs. Our design addresses this gap through three principles:

\vspace{0.5ex}
\noindent
\textbf{(1) Decomposition for multi-hop reasoning.} Complex queries often involve chained information needs where intermediate answers serve as inputs to subsequent questions. By decomposing queries into ordered sub-queries with explicit latent variables, we enable step-by-step reasoning where each step can be independently matched and verified.

\noindent
\textbf{(2) Taxonomy-guided matching.} Pure semantic similarity can be misled by lexical overlap or topic drift. By introducing hierarchical entity typing, we add a structural constraint that ensures type-level compatibility between query requirements and document content. This acts as a coarse filter that complements fine-grained semantic matching.

\noindent
\textbf{(3) Structured reasoning with evidence selection.} Rather than selecting all context at once, we iteratively resolve latent variables and refine context selection. This allows later reasoning steps to benefit from earlier resolutions, mimicking how humans would approach multi-hop questions by first answering prerequisite sub-questions.

\subsection{Overview}

We propose \textbf{TaSR-RAG} (\textbf{Ta}xonomy-guided \textbf{S}tructured \textbf{R}easoning for RAG), a multi-hop retrieval-augmented framework that treats taxonomy-guided evidence selection as an explicit reasoning process.
Given a complex query $q$, we (i) convert retrieved documents into relational triples and their corresponding taxonomy-typed triples, (ii) decompose the query into ordered sub-queries (triples with latent variables) and construct their typed counterparts, and (iii) traverse sub-queries from left to right, repeatedly reranking candidate documents with hybrid matching and using an LLM to resolve latent variables.

Let $\mathcal{D}$ denote the document collection, and let $\mathcal{T} = \{\mathcal{T}^{(1)}, \mathcal{T}^{(2)}\}$ denote a two-level taxonomy. $\mathcal{T}^{(1)} = \{c_1^{(1)}, \ldots, c_L^{(1)}\}$ are first-level classes and $\mathcal{T}^{(2)}(c^{(1)}) = \{c_1^{(2)}, \ldots, c_M^{(2)}\}$ are second-level classes under $c^{(1)}$. We build $\mathcal{T}$ based on Schema.org \footnote{\url{Schema.org}} and adopt a two-level hierarchy to balance coverage and discriminability.

The overall pipeline consists of the following stages:
\leftmargini=12pt
\vspace{-0.4\baselineskip}
\begin{enumerate}
    \setlength{\topsep}{0pt}
    \setlength{\partopsep}{0pt}
    \setlength{\itemsep}{0pt}
    \setlength{\parsep}{0pt}
  \item \textbf{Initial Retrieval}: retrieve top-$K_0$ candidate documents via dense retrieval.
  \item \textbf{Triple Extraction and Typing (Document-level)}: for each retrieved document, extract relational triples $\mathcal{R}_d$ and construct typed triples $\tilde{\mathcal{R}}_d$ by applying taxonomy-guided entity typing to the head and tail entities.
  \item \textbf{Query Decomposition and Typing (Query-level)}: decompose $q$ into an ordered list of sub-queries $\{s_i\}_{i=1}^N$ (triples with latent variables) and construct their typed forms $\{\tilde{s}_i\}_{i=1}^N$.
  \item \textbf{Structured Reasoning with Entity Binding}: for $i=1\ldots N$, resolve already-bound variables in $s_i$, rerank/filter documents using hybrid matching over $(s_i, \tilde{s}_i)$ against $(\mathcal{R}_d,\allowbreak\,\tilde{\mathcal{R}}_d)$, prompt the LLM to answer the current sub-query, and bind its latent variable for subsequent steps. The final answer is the output of the last sub-query.
\end{enumerate}


\noindent\textbf{Example~1.0} (\textit{Running example})
We use query $Q$:
``\textit{Which company originally developed the open-source relational database used in NASA/JPL's Science Activity Planner for the Mars Rover?}" as a running example throughout this section, illustrating the end-to-end pipeline at each stage (retrieval, Document-level triple construction, query decomposition, hybrid matching, and entity binding).

\subsection{Initial Retrieval}

Given query $q$, we first retrieve top-$K_0$ candidate documents via dense retrieval. This step is shared by all methods and serves as a common preprocessing stage for fair comparison:
\begin{equation}
  \mathcal{D}_{\text{cand}} = \underset{d \in \mathcal{D}}{\text{Top-}K_0} \cos\bigl(f_{\text{enc}}(q), f_{\text{enc}}(d)\bigr)
\end{equation}

\noindent\textbf{Example~1.1 (Fig.~\ref{fig:intro})}
For query Q in Ex.\ 1.0, dense retrieval returns candidate documents that mention Science Activity Planner  and database-related evidence (e.g., a sentence stating that the system uses the MySQL database). These retrieved documents form $\mathcal{D}_{\text{cand}}$ and will be converted into triples in the next stage.

\subsection{Triple Extraction and Typing (Document-level)}

\subsubsection{Triple Extraction}

After Example~1.1, we obtain a candidate document set $\mathcal{D}_{\text{cand}}$ that is likely to contain the multi-hop evidence needed by the query.
We now transform each candidate document into a compact, comparable set of relational triples.

\paragraph{Triple Extraction.}
For each candidate document $d \in \mathcal{D}_{\text{cand}}$, we employ an LLM to extract relational triples:
\begin{equation}
  \mathcal{R}_d = \text{LLM}_{\text{extract}}(d) = \{(h, r, t)_1, \ldots, (h, r, t)_{|\mathcal{R}_d|}\}
\end{equation}

\noindent\textbf{Example~1.2 (Fig.~\ref{fig:intro})}
From a retrieved document that states that \textit{Science Activity Planner} uses \textit{MySQL database}, the extracted triples may include:
\begin{align}
  (\text{Science Activity Planner}, \texttt{uses}, \text{MySQL database}), \\
  (\text{MySQL AB}, \texttt{developed}, \text{MySQL database}).
\end{align}

\paragraph{Triple Typing.}
After extracting $\mathcal{R}_d$, we assign hierarchical types to the head and tail entities (described in the next subsection) and obtain typed triples:
\begin{equation}
  \tilde{\mathcal{R}}_d = \bigl\{(\tau(h), r, \tau(t)) : (h, r, t) \in \mathcal{R}_d\bigr\}
\end{equation}
Note that relations $r$ retain their surface forms without typing, as their typing (i.e., generation) will lose its sharpness, confirmed by our experiments.

A natural question is why not extract triples from all documents beforehand. We adopt query-time extraction because incorporating the query during extraction significantly improves the relevance and accuracy of extracted triples. The query provides contextual guidance that helps the LLM focus on pertinent relations rather than exhaustively extracting all possible triples. We empirically compare this design with query-agnostic pre-extraction in our experiments (Section~\ref{sec:ablation}).

When extracting relations, we instruct the LLM to focus on relations that best characterize the relationship between two entities, avoiding both overly specific predicates and overly generic ones. This ensures that extracted relations are informative for matching while maintaining sufficient generality for cross-document comparison.

\subsubsection{Taxonomy-Guided Entity Typing}

We adopt a simple two-stage entity typing procedure that uses an LLM to directly select taxonomy labels.
For each entity $e$ (from extracted triples or query sub-queries), we assign a hierarchical type $\tau(e) = (c^{(1)}, c^{(2)})$ in a coarse-to-fine manner:

\vspace{0.5ex}
\noindent \textit{Stage 1: First-Level LLM Selection.}
The LLM selects the best first-level class from $\mathcal{T}^{(1)}$:
\begin{equation}
  c^{(1)} = \text{LLM}_{\text{select}}\bigl(\mathcal{T}^{(1)}, e\bigr)
\end{equation}

\vspace{0.5ex}
\noindent \textit{Stage 2: Second-Level LLM Selection.}
Conditioned on $c^{(1)}$, the LLM selects the best second-level class from its children:
\begin{equation}
  c^{(2)} = \text{LLM}_{\text{select}}\bigl(\mathcal{T}^{(2)}(c^{(1)}), e\bigr)
\end{equation}

The final hierarchical type is $\tau(e) = (c^{(1)}, c^{(2)})$, where $c^{(1)}$ is the parent of $c^{(2)}$.
For query-level latent variables (e.g., ?Olympics), the type is inferred from the variable semantics during query decomposition and used as a constraint in later matching.

\vspace{0.5ex}
\noindent\textbf{Example~1.3 (Fig.~\ref{fig:intro})}
Continuing from Ex.~1.2, we assign each entity a two-level taxonomy type $\tau(e)=(c^{(1)},c^{(2)})$.
These types yield the corresponding typed triples used in downstream matching.

\subsection{Query Decomposition and Typing (Query-level)}

After constructing (typed) triples for the retrieved documents (Examples~1.1--1.3), we decompose the complex multi-hop query $q$ into an ordered sequence of sub-queries that explicitly exposes the intermediate unknowns and guides step-wise reranking.
Given a complex multi-hop query $q$, we employ an LLM to decompose it into an ordered sequence of sub-queries:
\begin{equation}
  \{s_1, s_2, \ldots, s_N\} = \text{LLM}_{\text{decomp}}(q)
\end{equation}
where each sub-query $s_i$ is represented as a relational triple:
\begin{equation}
  s_i = (h_i, r_i, t_i)
\end{equation}
Here $h_i$ denotes the head entity, $r_i$ denotes the relation, and $t_i$ denotes the tail entity. Crucially, either $h_i$ or $t_i$ may be a \textit{latent variable} (denoted with prefix ``?''), representing an unknown entity to be resolved through reasoning.

Because the downstream matching is performed over both raw triples and taxonomy-typed triples, our decomposition prompt also enforces query-level typing semantics: for each explicit entity and latent variable, the LLM infers a coarse type description that is mapped into the two-level taxonomy, yielding $\tau(h_i)$ and $\tau(t_i)$ (or type constraints for variables). This makes query decomposition the place where query-level triple formation and (lightweight) typing are performed, complementing the document-level triple extraction and typing above.

\noindent\textbf{Example~1.4 (decomposition) (Fig.~\ref{fig:intro})}
For the query in Section~\ref{sec:analysis}, the LLM produces two sub-queries with latent variables:
\begin{align}
  s_1 &= (\text{Science Activity Planner}, \texttt{uses}, \text{?Database}), \\
  s_2 &= (\text{?Database}, \texttt{developed\_by}, \text{?Company}).
\end{align}


\noindent\textbf{Example~1.5 (Query-level typed triples) (Fig.~\ref{fig:intro})}
We then apply taxonomy-guided typing to both explicit entities and latent variables to obtain typed sub-queries:
\begin{align}
  \tilde{s}_1 &= (\textsc{WORK/System}, \texttt{uses}, \textsc{WORK/Software}), \\
  \tilde{s}_2 &= (\textsc{WORK/Software}, \texttt{developed\_by}, \textsc{Organization/Company}).
\end{align}
These typed sub-queries are used together with the raw sub-queries in document reranking.

\paragraph{Query-level Typed Triples.}
After obtaining $\{s_i\}_{i=1}^N$, we apply the same taxonomy-guided entity typing procedure (Section~\ref{sec:ablation}) to the head and tail entities in each sub-query and obtain the typed form
\begin{equation}
  \tilde{s}_i = (\tau(h_i), r_i, \tau(t_i)).
\end{equation}
These typed sub-queries provide type-level constraints for downstream matching and are used in our taxonomy-guided document scoring module.

The decomposition is implemented via prompting, where the LLM is instructed to identify the reasoning steps required to answer the query and express each step as a relational triple with explicit latent variables. We provide an error analysis of decomposition quality in Section \ref{sec:error_analysis}.

\subsection{Structured Reasoning with Entity Binding}

\subsubsection{Hybrid Triple Matching for Document Reranking}

For each resolved sub-query $s_i'$ and its typed form $\tilde{s}_i'$, we score each candidate document $d$ by matching the query triple against the document's extracted triples and matching the typed query triple against the document's typed triples.
We then combine the two signals into a single score used to rerank documents for the current reasoning step.

\noindent\textbf{Example~1.6 (hybrid matching) (Fig.~\ref{fig:intro})}
For the first-hop sub-query $s_1$, the semantic score favors document triples that are close in meaning to the query.
The structural score enforces type compatibility between matched entities.

\paragraph{Structural Score.}
The structural score measures type-level compatibility between the query triple and document triples. For a query-document triple pair, we compute:
\begin{equation}
  S_{\text{type}}(\tau_q, \tau_d) = w^{(1)} \cdot \mathbb{1}[c_q^{(1)} = c_d^{(1)}] + w^{(2)} \cdot \mathbb{1}[c_q^{(2)} = c_d^{(2)}]
\end{equation}
where $\tau_q = (c_q^{(1)}, c_q^{(2)})$ and $\tau_d = (c_d^{(1)}, c_d^{(2)})$ are hierarchical types, and $w^{(1)} + w^{(2)} = 1$ control the relative importance of each level.

The structural score for a triple pair is:
\begin{equation}
  S_{\text{struct}}(\tilde{s}_i, \tilde{\rho}) = w_h \cdot S_{\text{type}}(\tau(h_i), \tau(h)) + w_t \cdot S_{\text{type}}(\tau(t_i), \tau(t))
\end{equation}
where $\tilde{\rho} = (\tau(h), r, \tau(t)) \in \tilde{\mathcal{R}}_d$ and $w_h + w_t = 1$.

\paragraph{Semantic Score.}
The semantic score captures embedding-level similarity between raw triples. We encode each component separately with role-specific prefixes:
\begin{equation}
  \mathbf{v}_h = f_{\text{enc}}(\texttt{"S: "} \| h), \quad \mathbf{v}_r = f_{\text{enc}}(\texttt{"P: "} \| r), \quad \mathbf{v}_t = f_{\text{enc}}(\texttt{"O: "} \| t)
\end{equation}
The semantic score between query triple $s_i = (h_i, r_i, t_i)$ and document triple $\rho = (h, r, t)$ is:
\begin{equation}
  S_{\text{sem}}(s_i, \rho) = \lambda_h \cos(\mathbf{v}_{h_i}, \mathbf{v}_h) + \lambda_r \cos(\mathbf{v}_{r_i}, \mathbf{v}_r) + \lambda_t \cos(\mathbf{v}_{t_i}, \mathbf{v}_t)
\end{equation}
where $\lambda_h + \lambda_r + \lambda_t = 1$.

\paragraph{Triple-Level Score.}
The overall score for a query-document triple pair combines structural and semantic components:
\begin{equation}
  S_{\text{triple}}(s_i, \rho) = \alpha \cdot S_{\text{struct}}(\tilde{s}_i, \tilde{\rho}) + (1 - \alpha) \cdot S_{\text{sem}}(s_i, \rho)
\end{equation}

\paragraph{Document-Level Aggregation.}
To obtain a document-level score, we first compute the best matching score for each query triple within the document:
\begin{equation}
  S_{\text{best}}(s_i, d) = \max_{\rho \in \mathcal{R}_d} S_{\text{triple}}(s_i, \rho)
\end{equation}
The document score aggregates across all query triples using a mixture of max and mean:
\begin{equation}
  S(d) = \gamma \cdot \max_{i} S_{\text{best}}(s_i, d) + (1 - \gamma) \cdot \frac{1}{|\mathcal{S}_t|} \sum_{s \in \mathcal{S}_t} S_{\text{best}}(s, d)
\end{equation}
where $\mathcal{S}_t$ denotes the top-$t$ sub-queries ranked by their best matching scores.

\paragraph{Threshold Filtering and Reranking.}
Documents with scores below a threshold $\theta$ are filtered out, and the remaining documents are reranked by $S(d)$:
\begin{equation}
  \mathcal{D}_i = \text{Rank}\bigl(\{d \in \mathcal{D}_{\text{cand}} : S(d) \geq \theta\};\, S(d)\bigr)
\end{equation}
This reranked list $\mathcal{D}_i$ is then passed to the LLM to answer the current sub-query and resolve its latent variables.

\subsubsection{Entity Binding and Sub-query Execution}

We maintain an entity binding table $\mathcal{B}$ that maps latent variables to their resolved values, and iteratively resolve latent variables while reranking evidence at each step.

\paragraph{Initialization.}
\begin{equation}
  \mathcal{B}^{(0)} = \emptyset
\end{equation}

\paragraph{Iterative Resolution.}
For each sub-query $s_i$ in order:

\noindent\textit{Step 1: Variable Substitution.}
\begin{equation}
  s_i' = \text{Resolve}(s_i, \mathcal{B}^{(i-1)})
\end{equation}
where $\text{Resolve}(\cdot)$ substitutes any latent variable in $s_i$ with its binding from $\mathcal{B}^{(i-1)}$ if available.

\noindent\textit{Step 2: Evidence Selection (Filter + Rerank).}
Compute document scores using $s_i'$ (and $\tilde{s}_i'$) and obtain a filtered, reranked document list:
\begin{equation}
  \mathcal{D}_i = \text{Rank}\bigl(\{d \in \mathcal{D}_{\text{cand}} : S(d \mid s_i') \geq \theta\};\, S(d \mid s_i')\bigr)
\end{equation}

\noindent\textit{Step 3: Sub-query Answering.}
Given the reranked documents, the LLM answers the current sub-query and outputs the value of its latent variable (if any):
\begin{equation}
  \hat{o}_i = \text{LLM}_{\text{answer}}(s_i', \mathcal{D}_i)
\end{equation}

\noindent\textit{Step 4: Entity Binding Update.}
\begin{equation}
  \mathcal{B}^{(i)} = \mathcal{B}^{(i-1)} \cup \{v_i \mapsto \hat{o}_i\}
\end{equation}
where $v_i$ is the latent variable in $s_i$ (if any).

After processing all sub-queries, the answer to the original question is taken to be the output of the final sub-query, i.e., $\hat{a} = \hat{o}_N$, together with its supporting evidence from the reranked documents in the last step.

\noindent\textbf{Example~1.7 (entity binding) (Fig.~\ref{fig:intro})}
In the running example, the first step resolves
\begin{equation*}
  \text{?Database} = \text{MySQL database}.
\end{equation*}
We then substitute this binding into $s_2$ and rerank documents again to resolve
\begin{equation*}
  \text{?Company} = \text{MySQL AB}.
\end{equation*}

\begin{table*}[htbp]
  \centering
  \caption{Performance comparison (EM / F1) of different methods on Qwen2.5-7B-Instruct and Qwen2.5-72B-Instruct. The best results are in \textbf{bold} and the second best are \underline{underlined}.}
  \label{tab:performance}
  \resizebox{0.98\textwidth}{!}{%
    \begin{tabular}{lcccccccccccccccc}
      \toprule
      \multirow{3}{*}{\textbf{Methods}}
      & \multicolumn{6}{c}{\textbf{General QA}}
      & \multicolumn{8}{c}{\textbf{Multi-Hop QA}}
      & \multirow{3}{*}{\textbf{Avg. EM}}
      & \multirow{3}{*}{\textbf{Avg. F1}} \\
      \cmidrule(lr){2-7} \cmidrule(lr){8-15}
      & \multicolumn{2}{c}{NQ}
      & \multicolumn{2}{c}{TriviaQA}
      & \multicolumn{2}{c}{PopQA}
      & \multicolumn{2}{c}{HotpotQA}
      & \multicolumn{2}{c}{2WikimQA}
      & \multicolumn{2}{c}{Musique}
      & \multicolumn{2}{c}{Bamboogle}
      & & \\
      & EM & F1 & EM & F1 & EM & F1
      & EM & F1 & EM & F1 & EM & F1 & EM & F1
      & & \\
      \midrule
      \multicolumn{17}{c}{\cellcolor{gray!20}\textbf{Qwen2.5-72B-Instruct}} \\
      Direct Inference
      & 19.9 & 28.3 & 45.4 & 53.1 & 19.7 & 24.4
      & 20.6 & 29.0 & 15.3 & 19.6 & 5.5 & 10.3 & 17.6 & 27.8
      & 20.6 & 27.5 \\
      CoT
      & 23.8 & 33.8 & 51.6 & 60.4 & 24.4 & 30.3
      & 26.4 & 37.2 & 20.5 & 26.2 & 9.9 & 18.6 & 19.6 & 31.0
      & 25.2 & 33.9 \\
      RAG
      & 34.2 & \textbf{48.6} & 53.4 & 62.5 & 33.2 & 41.2
      & 32.8 & 46.2 & 22.7 & 29.1 & 10.6 & 19.9 & 21.6 & 34.1
      & 29.7 & 40.2 \\
      IRCoT
      & 23.3 & 33.1 & \textbf{71.9} & \textbf{84.2} & \textbf{52.0} & \textbf{64.5}
      & 21.9 & 30.8 & 25.1 & 32.1 & \underline{13.2} & \textbf{24.8} & 23.2 & 36.6
      & 32.9 & 43.7 \\
      GraphRAG
      & 26.6 & 38.5 & 61.7 & 70.9 & 36.7 & 45.8
      & 31.2 & 42.7 & 30.9 & 38.5 & 9.2 & 17.9 & 6.4 & 15.8
      & 29.0 & 38.6 \\
      HippoRAG
      & \underline{36.5} & 48.2 & 61.9 & \underline{71.9} & \underline{41.1} & \underline{49.7}
      & \underline{37.2} & \textbf{48.5} & \textbf{55.5} & \underline{63.6} & 11.7 & 19.8 & 29.6 & 35.9
      & \underline{39.1} & \underline{48.2} \\
      HyperGraphRAG
      & 25.7 & 37.3 & 57.4 & 68.1 & 36.0 & 45.9
      & 29.6 & \underline{42.8} & 42.1 & 51.5 & 9.0 & 17.4 & 19.2 & 27.0
      & 31.3 & 41.4 \\
      StructRAG
      & 27.9 & 40.6 & 60.8 & \underline{71.9} & 36.5 & 45.0
      & 27.5 & 41.8 & 24.5 & 37.2 & 8.8 & 17.2 & \underline{44.8} & \textbf{54.5}
      & 33.0 & 44.0 \\
      \textbf{Ours}
      & \textbf{40.6} & \underline{48.3} & \underline{64.5} & 69.9 & 37.7 & 43.2
      & \textbf{38.7} & \textbf{48.5} & \underline{51.8} & \textbf{66.2} & \textbf{18.3} & \underline{24.4} & \textbf{45.6} & \underline{49.1}
      & \textbf{42.5} & \textbf{49.9} \\
      \midrule
      \multicolumn{17}{c}{\cellcolor{gray!20}\textbf{Qwen2.5-7B-Instruct}} \\
      Direct Inference
      & 13.4 & 23.2 & 40.8 & 49.7 & 14.0 & 16.4
      & 18.3 & 27.1 & 12.6 & 30.1 & 3.1 & 11.9 & 12.0 & 17.2
      & 16.3 & 25.1 \\
      CoT
      & 4.8 & 19.3 & 18.5 & 49.7 & 5.4 & 16.4
      & 9.2 & 13.0 & 10.8 & 15.1 & 2.2 & 5.1 & \underline{23.2} & 20.8
      & 10.6 & 19.9 \\
      RAG
      & \underline{32.9} & \underline{40.0} & 35.5 & \textbf{67.2} & 19.2 & \underline{44.3}
      & 19.9 & \underline{41.9} & 13.5 & \underline{47.7} & 5.8 & 13.0 & 20.8 & 22.4
      & 21.1 & \underline{39.5} \\
      IRCoT
      & 22.4 & 34.7 & 47.8 & \underline{65.9} & \underline{30.1} & \textbf{60.2}
      & 13.3 & 20.2 & 14.9 & 21.3 & \underline{7.2} & \underline{20.7} & 22.4 & \textbf{32.1}
      & \underline{22.6} & 36.4 \\
      GraphRAG
      & 22.0 & 21.4 & 35.9 & 48.0 & 12.8 & 27.4
      & \underline{22.1} & 33.6 & \underline{30.5} & 37.8 & 4.1 & 11.8 & 8.0 & 14.9
      & 17.0 & 27.9 \\
      HippoRAG
      & 18.2 & 25.1 & 37.0 & 51.0 & 26.8 & 37.0
      & 18.2 & 32.2 & 19.7 & 38.2 & 4.4 & 12.8 & 15.2 & 23.7
      & 18.8 & 31.4 \\
      HypergraphRAG
      & 1.3 & 2.7 & 3.8 & 6.2 & 0.8 & 1.6
      & 4.4 & 6.2 & 1.6 & 1.9 & 0.4 & 1.2 & 3.2 & 3.9
      & 2.2 & 3.4 \\
      StructRAG
      & 20.8 & 32.2 & \underline{50.0} & 60.4 & 31.4 & 40.1
      & 18.5 & 29.8 & 16.7 & 23.9 & 4.1 & 10.7 & 13.6 & 24.5
      & 22.2 & 31.7 \\
      \textbf{Ours}
      & \textbf{38.2} & \textbf{45.2} & \textbf{55.1} & 60.5 & \textbf{36.5} & 43.9
      & \textbf{35.1} & \textbf{42.0} & \textbf{45.9} & \textbf{65.6} & \textbf{14.6} & \textbf{22.0} & \textbf{25.6} & \underline{28.6}
      & \textbf{35.9} & \textbf{44.0} \\ \bottomrule
    \end{tabular}
  }
\end{table*}

\section{Experiments}
To demonstrate the efficacy of our TaSR-RAG, we conduct extensive evaluations on a variety of open-domain and multi-hop Question Answering (QA) benchmarks.

\subsection{Datasets and Evaluation Metrics}
Our evaluation spans seven representative knowledge-intensive datasets, selected to cover a broad spectrum of reasoning challenges. For single-step retrieval, we utilize Natural Questions (NQ) \cite{nq}, TriviaQA \cite{triviaqa}, and PopQA \cite{popqa}. To investigate complex reasoning, we include several multi-hop datasets: HotpotQA \cite{hotpotqa}, 2WikiMultiHopQA \cite{2wikimqa}, MuSiQue \cite{musique}, and Bamboogle \cite{bamboogle}. Together, these benchmarks provide a rigorous testing ground for both basic retrieval-augmented generation and sophisticated multi-step reasoning.

We employ standard QA metrics, specifically Exact Match (EM) and F1 scores, as the primary performance indicators, following prior studies \cite{search-r1, structure-r1, search-o1}.


\subsection{Baselines}
We compare the performance of TaSR-RAG against two distinct groups of baseline models. The first category consists of Retrieval-free Methods: (1) Direct Inference, which relies solely on the generator's internal knowledge; (2) Chain-of-Thought (CoT) \cite{cot}, utilizing step-by-step prompting. The second category encompasses Retrieval-augmented Frameworks, with a particular focus on state-of-the-art graph-based approaches: (1) RAG \cite{replug}, the conventional dense retrieval-generation pipeline; (2) IRCoT \cite{IRCoT}, which interleaves retrieval with intermediate reasoning steps to iteratively gather multi-hop evidence; (3) GraphRAG \cite{graphrag}, which leverages community detection and global summarization over knowledge graphs to capture holistic dataset themes; (4) HippoRAG \cite{hipporag}, a neurobiologically inspired framework that employs Personalized PageRank to achieve long-term memory integration and multi-hop associative retrieval; (5) HyperGraphRAG \cite{hypergraphrag}, which models high-order relationships using hyperedges to better represent complex, multi-entity interactions; and (6) StructRAG \cite{structrag}, a structured retrieval-augmented approach that dynamically adapts the retrieval granularity and structural organization based on the complexity of the query.
These baselines encompass a wide array of prompting and fine-tuning strategies, providing a comprehensive context for our results.

\subsection{Main Results}

Table~\ref{tab:performance} reports Exact Match (EM) results on seven QA benchmarks using \texttt{Qwen2.5-72B-Instruct} and \texttt{Qwen2.5-7B-Instruct} as the underlying generators.
Across both model scales, \textsc{TaSR-RAG} achieves the best overall average performance, indicating that taxonomy-typed triples and sequential latent-variable binding improve evidence selection robustness.

\noindent\textbf{Qwen2.5-72B-Instruct.}
Our method achieves the highest average EM (42.5), substantially outperforming standard RAG (29.7) and other retrieval-augmented baselines.
At the dataset level, \textsc{TaSR-RAG} attains the best scores on NQ (40.6), HotpotQA (38.7), Musique (18.3), and Bamboogle (45.6), while remaining competitive on TriviaQA (64.5), PopQA (37.7), and 2WikimQA (51.8).
Notably, the strongest competing methods are highly dataset-dependent (e.g., IRCoT on TriviaQA and HippoRAG on 2WikimQA), whereas our approach provides consistently strong results across both general and multi-hop QA.

\noindent\textbf{Qwen2.5-7B-Instruct.}
The gains are even more pronounced in the smaller-model setting.
\textsc{TaSR-RAG} achieves the highest average EM (37.0), improving over standard RAG (21.1) by a large margin.
Moreover, it ranks first on all seven datasets, suggesting that explicit structure and step-by-step binding can effectively compensate for limited model capacity by guiding the model to focus on high-density, type-consistent evidence.

\begin{table}[t]
  \centering
  \caption{Ablation to disentangle the effects of Document-level and Query-level structuredization under the Qwen2.5-72B-Instruct setting.
    Doc-level Structure applies triple extraction and taxonomy typing to documents while keeping the original query unchanged.
  Query-level Structure enables query decomposition and typed sub-queries but disables document-level structural representations.}
  \setlength{\tabcolsep}{8pt}
  \resizebox{0.48\textwidth}{!}{%
    \begin{tabular}{lcccc}
      \toprule
      \textbf{Method} & \textbf{HotpotQA} & \textbf{2WikiMQA} & \textbf{Bamboogle} & \textbf{Avg.} \\
      \midrule
      RAG         & 32.8 & 22.7 & 21.6 & 25.7 \\
      Doc-level Structure    & 44.0 & 34.0 & 40.8 & 39.6 \\
      Query-level Structure    & 42.5 & 31.5 & 43.2 & 39.1 \\
      Full       & 46.2 & 35.6 & 45.1 & 42.3 \\
      \bottomrule
    \end{tabular}
  }
  \label{tab:ablation_query_doc_structure}
\end{table}

\subsection{Ablation Studies}
\label{sec:ablation}

To better understand the contribution of each component in TaSR-RAG, we conduct a series of ablation studies that systematically remove or modify key design choices. All ablation variants are evaluated under the same experimental settings and retrieval budgets for fair comparison. Results are reported using Exact Match (EM).

\subsection{Query-level vs. Document-level Structuredization}

Our framework introduces structure on both the document side and the query side, raising a natural question: which component contributes more to the overall performance gains?
To disentangle their individual effects, we conduct a targeted ablation study under the Qwen2.5-72B setting on three representative multi-hop benchmarks.
Specifically, we compare (i) a Doc-level variant, which converts retrieved documents into relational triples and taxonomy-typed triples while keeping the query as a single, undecomposed request, and (ii) a Query-level variant, which enables query decomposition into sequential, typed sub-queries but leaves documents in their original unstructured form.
As shown in Table~\ref{tab:ablation_query_doc_structure}, both variants substantially outperform the unstructured RAG baseline, indicating that structured representations on either side already provide strong inductive bias for multi-hop reasoning.
Interestingly, the two variants achieve comparable average performance (39.6 vs.\ 39.1 EM), but exhibit different dataset-specific strengths: Document-level structure is slightly more effective on HotpotQA and 2WikiMQA, while query-level structure yields stronger gains on Bamboogle, which involves more implicit multi-step inference.
The full model, which combines structured queries with structured documents, achieves the best results across all datasets, improving the average EM to 42.3.
Crucially, their combination is complementary rather than redundant, and both are necessary to fully realize the benefits of structured, sequential reasoning in retrieval-augmented generation.
Moreover, the ablation helps pinpoint where each form of structure matters most: query-level decomposition primarily stabilizes hop ordering and reduces reasoning drift, while document-level tripleization improves evidence precision by filtering distractor spans.
This motivates our full design, which aligns structured queries with structured documents to maximize both recall of relevant hops and faithfulness of the final answer.

The full TaSR-RAG consistently outperforms both variants, with especially pronounced improvements on multi-hop datasets.
While No-Seq suffers from limited reasoning depth, Seq-NoBind exhibits clear degradation, highlighting the importance of maintaining consistent entity bindings across reasoning steps.
These results demonstrate that sequential reasoning alone is insufficient without explicit variable binding.

\subsection{Impact of Taxonomy-Guided Entity Typing}

\begin{table}[t]
  \centering
  \caption{Effect of taxonomy granularity on multi-hop QA performance under the Qwen2.5-72B-Instruct setting.}
  \setlength{\tabcolsep}{8pt}
  \resizebox{0.48\textwidth}{!}{%
    \begin{tabular}{lcccc}
      \toprule
      \textbf{Method} & \textbf{HotpotQA} & \textbf{2WikiMQA} & \textbf{Bamboogle} & \textbf{Avg.} \\
      \midrule
      No Typing              & 32.8 & 22.7 & 21.6 & 25.7 \\
      Top-level Taxonomy     & 43.0 & 34.5 & 33.6 & 37.0 \\
      Top-two level Taxonomy & 46.2 & 35.6 & 45.1 & 42.3 \\
      Top-three level Taxonomy & 42.0 & 34.1 & 36.0 & 37.4 \\
      \bottomrule
    \end{tabular}
  }
  \label{tab:ablation_taxonomy_granularity}
\end{table}

\begin{table}[t]
  \centering
  \caption{Ablation on matching strategies under the Qwen2.5-72B setting.
  Semantic-only matching relies solely on triple embedding similarity, structural-only matching uses typed triple consistency, and hybrid matching combines both signals.}
  \setlength{\tabcolsep}{8pt}
  \resizebox{0.48\textwidth}{!}{%
    \begin{tabular}{lcccc}
      \toprule
      \textbf{Matching Strategy} & \textbf{HotpotQA} & \textbf{2WikiMQA} & \textbf{Bamboogle} & \textbf{Avg.} \\
      \midrule
      Semantic-only Matching     & 43.5 & 23.5 & 43.2 & 36.7 \\
      Structural-only Matching   & 44.0 & 33.5 & 44.8 & 40.8 \\
      Hybrid Matching (Full)     & 46.2 & 35.6 & 45.1 & 42.3 \\
      \bottomrule
    \end{tabular}
  }
  \label{tab:ablation_matching_strategy}
\end{table}

We next investigate how the granularity of entity typing affects structured retrieval and reasoning.
Using the same pipeline and retrieval budget, we vary the depth of the taxonomy used for typing entities on both the query and document sides.
Here, \textbf{Top-level Taxonomy} uses only the top-level taxonomy (L1), \textbf{Top-two level Taxonomy} uses the top two levels (L1+L2), and \textbf{Top-three level Taxonomy} uses three levels (L1+L2+L3).
As shown in Table~\ref{tab:ablation_taxonomy_granularity}, introducing even Top-level typing leads to a substantial improvement over the untyped variant, confirming that explicit type constraints are essential for reducing retrieval noise and entity conflation.
Moving from Top-level to Top-two level yields further gains across all datasets, with especially pronounced improvements on Bamboogle, suggesting that mid-level semantic distinctions provide stronger guidance for multi-step inference.
However, increasing the granularity to Top-three level results in consistent performance degradation, despite using more detailed type information.
We attribute this drop to increased sparsity and error propagation: overly fine-grained types reduce matching recall and amplify the impact of occasional typing errors, which in turn destabilize downstream reasoning.

\subsection{Effect of Matching Strategy}

We further analyze the contribution of the proposed hybrid matching mechanism by comparing it against single-mode alternatives.
Specifically, we evaluate semantic-only matching, which scores documents solely based on embedding similarity between query and document triples, and structural-only matching, which relies exclusively on taxonomy-typed triple consistency.
As shown in Table~\ref{tab:ablation_matching_strategy}, semantic-only matching performs well on datasets with strong lexical or semantic overlap, but suffers significantly on 2WikiMQA, indicating vulnerability to topic drift and insufficient structural constraints during multi-hop retrieval.
In contrast, structural-only matching yields higher and more stable performance, confirming that typed structural constraints effectively reduce noise and entity conflation; however, its performance remains limited by reduced recall when exact structural matches are sparse.
The hybrid strategy consistently outperforms both alternatives across all datasets, achieving the highest average EM.
These results demonstrate that semantic similarity and structural consistency provide complementary signals: semantic matching ensures coverage and recall, while structural matching enforces type-level precision.

\subsection{Analysis}
\label{sec:analysis}


Beyond aggregate performance, we provide further analyses to shed light on the behavior and robustness of \textsc{TaSR-RAG}.

\subsection{Error Breakdown Analysis.}
\label{sec:error_analysis}
We manually inspect a subset of error cases where the ground-truth evidence is present in the retrieved pool, and conduct an error analysis to isolate failures introduced by our pipeline. Following our two-tier taxonomy, we group errors into (i) \textbf{\textsc{Matching Errors}} arising from typed triple matching and reranking, and (ii) \textbf{\textsc{Generation Errors}} arising from sequential reasoning over the selected context.

\begin{figure}[t]
  \centering

  %
  %

  \begin{subfigure}{0.4\linewidth}
    \centering
    \includegraphics[width=\linewidth]{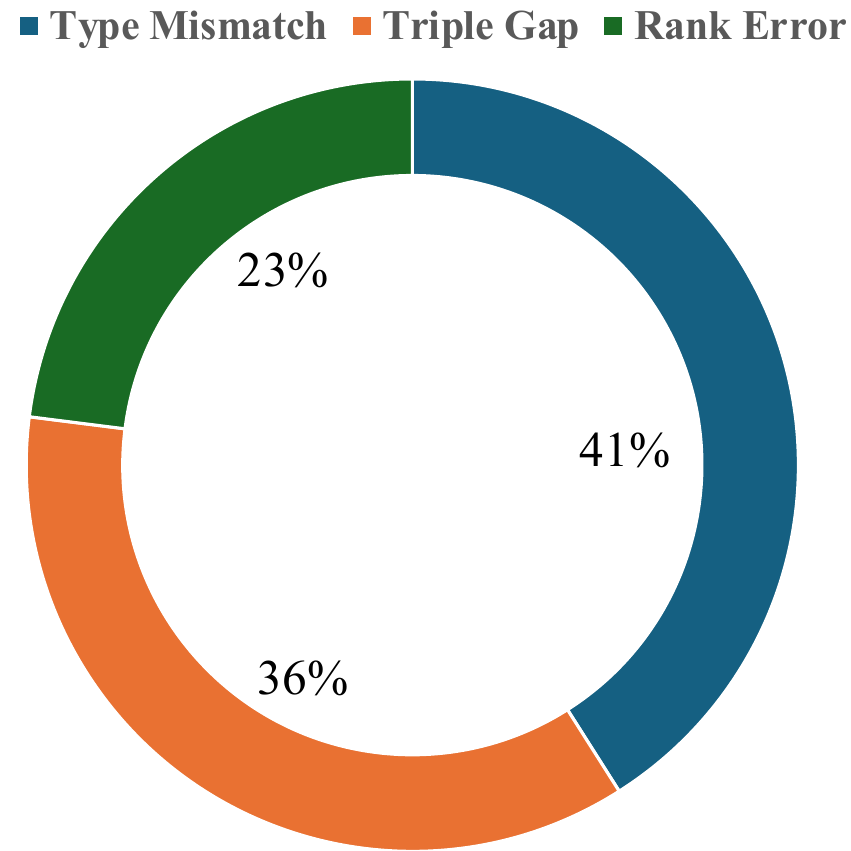}
  \end{subfigure}
  \hfill
  \begin{subfigure}{0.49\linewidth}
    \centering
    \includegraphics[width=\linewidth]{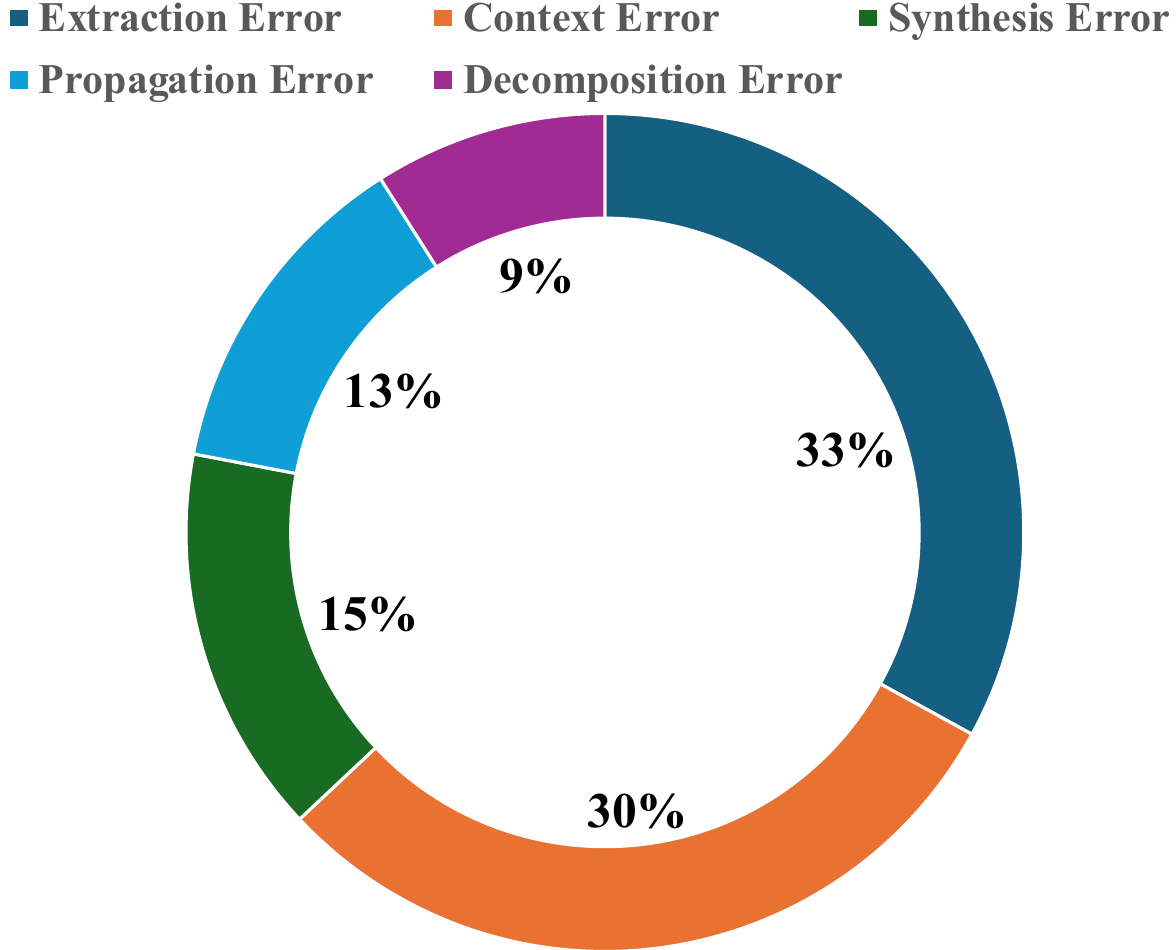}
  \end{subfigure}

  \caption{Error distribution in our manual analysis. The left figure illustrates Matching Errors and right figure illustrates Generation Errors.}
  \label{fig:error_tier}
\end{figure}

As shown in Figure~\ref{fig:error_tier}, most failures (64.5\%) occur after retrieval and reranking, indicating that the dominant bottleneck is reasoning and answer construction rather than document ranking.

\noindent\textbf{Generation errors.}
Within sequential reasoning, the most frequent issues are \textbf{Extraction Error} (33\%) and \textbf{Context Error} (30\%), where relevant documents are highly ranked but the LLM either fails to extract the exact answer string (format/normalization) or fails to derive the correct step result from the provided evidence. This suggests that, once relevant evidence is retrieved, the dominant bottleneck shifts to faithful utilization of the selected context: extraction failures are often triggered by aliasing/normalization (e.g., abbreviations, date formats) and by answers expressed implicitly rather than as a single span, while context errors indicate that hop-level evidence can still be noisy or incomplete, causing the model to follow a locally plausible but globally incorrect chain. We also observe \textbf{Synthesis Error} (15\%) where intermediate steps appear consistent but the final answer becomes inconsistent, and \textbf{Propagation Error} (13\%) caused by incorrect variable substitution across sub-queries. Finally, \textbf{Decomposition Error} (10\%) suggests that imperfect sub-query formation remains a non-negligible source of downstream failures.

\begin{figure}[t]
  \centering
  \begin{subfigure}[t]{0.9\linewidth}
    \centering
    \includegraphics[width=\linewidth]{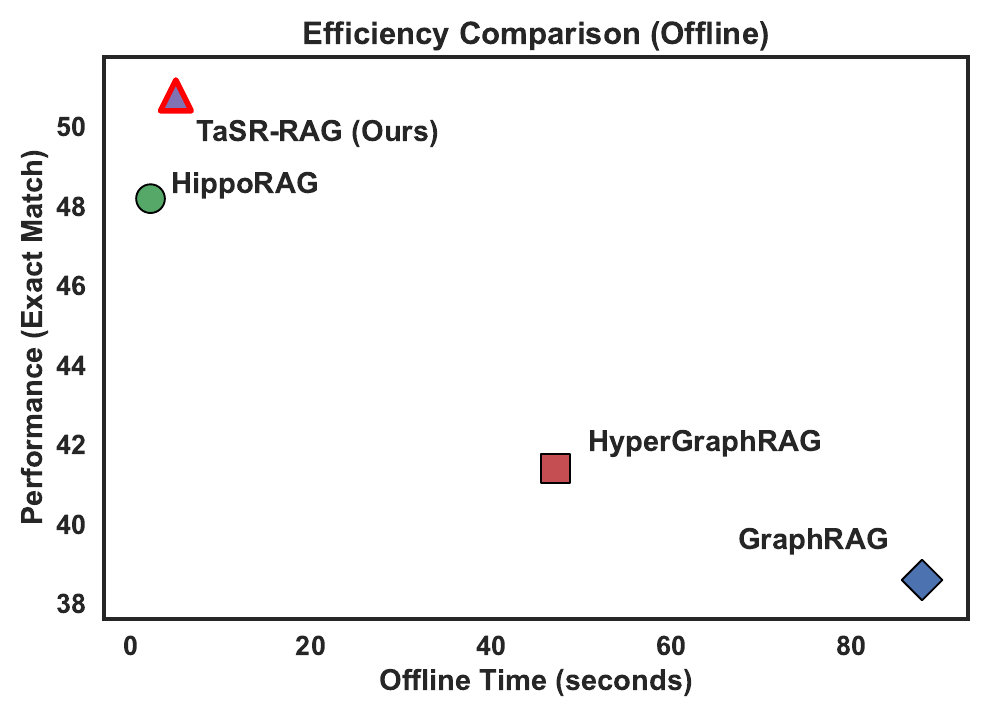}
    \label{fig:efficiency_offline}
    \caption{Offline efficiency.}
  \end{subfigure}

  \vspace{0.8em}

  \begin{subfigure}[t]{0.9\linewidth}
    \centering
    \includegraphics[width=\linewidth]{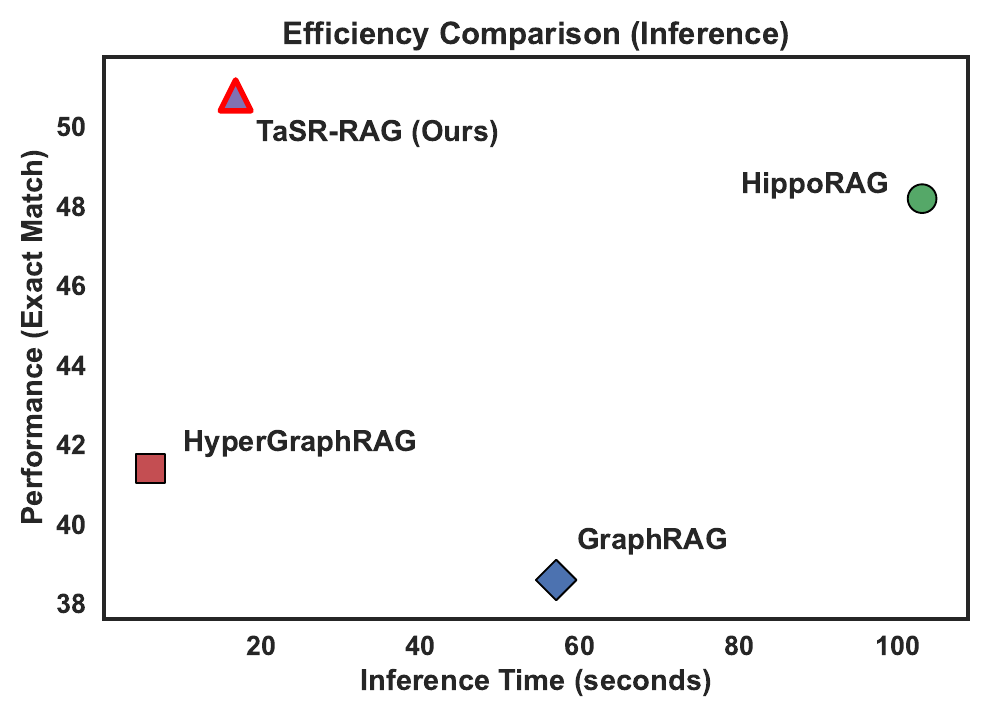}
    \label{fig:efficiency_inference}
    \caption{Inference efficiency.}
  \end{subfigure}

  \caption{
    Performance--efficiency trade-off under offline preprocessing and inference-time execution.
    \textsc{TaSR-RAG} (ours) achieves the best overall balance, avoiding the high offline cost of graph-based methods while maintaining low inference-time overhead.
  }
  \label{fig:efficiency}
\end{figure}

\noindent\textbf{Matching errors.}
Among typed triple matching failures, \textbf{Type Mismatch} (41\%) and \textbf{Triple Gap} (36\%) dominate. This highlights a key trade-off in our structured pipeline: typing constraints improve precision, but they also amplify upstream typing mistakes and brittleness in triple extraction/coverage. The remaining \textbf{Rank Error} (23\%) suggests that even when types and triples are adequate, the hybrid (semantic+structural) scoring can still mis-rank evidence.


\subsection{Efficiency}
\label{sec:efficiency}

We analyze the computational efficiency of different RAG frameworks by separating the cost into two stages: offline preprocessing and inference-time execution.
Figure~\ref{fig:efficiency} reports the trade-off between performance and time cost for \textsc{GraphRAG}, \textsc{HippoRAG}, \textsc{HyperGraphRAG}, and \textsc{TaSR-RAG}.

\paragraph{Offline Efficiency.}
Offline preprocessing exhibits substantial variance across methods.
Graph-based approaches such as \textsc{GraphRAG} and \textsc{HyperGraphRAG} incur high offline costs due to expensive graph construction and indexing, with \textsc{GraphRAG} being particularly costly.
In contrast, \textsc{HippoRAG} achieves very low preprocessing time, while \textsc{TaSR-RAG} requires only a modest offline cost despite incorporating structured retrieval and matching.
Importantly, \textsc{TaSR-RAG} achieves the highest performance among all methods while maintaining an offline time that is an order of magnitude lower than heavy graph-based baselines.

\paragraph{Inference Efficiency.}
Inference-time efficiency reveals a different trade-off.
While \textsc{HyperGraphRAG} achieves fast inference by relying on precomputed structures, \textsc{HippoRAG} suffers from significant inference overhead.
\textsc{GraphRAG} also incurs non-trivial inference cost.
In contrast, \textsc{TaSR-RAG} maintains consistently low inference time while delivering the best performance, demonstrating that its structured reasoning and retrieval design does not introduce prohibitive online overhead.

Across both offline and inference settings, \textsc{TaSR-RAG} occupies a favorable position on the performance--efficiency frontier.
It avoids the extreme offline cost of graph-heavy methods and the high inference latency of retrieval-intensive approaches, achieving a balanced and practical trade-off without sacrificing accuracy.

\section{Conclusion}
\label{sec:conclusion}

We presented \textsc{TaSR-RAG}, a taxonomy-guided sequential reasoning framework for multi-hop context selection in retrieval-augmented generation. By representing queries and documents as relational triples, imposing lightweight two-level type constraints, and combining semantic and structural signals for step-wise reranking, \textsc{TaSR-RAG} selects denser and more relevant evidence than standard chunk-based retrieval. Crucially, \textsc{TaSR-RAG} maintains an explicit entity binding table to resolve latent variables across reasoning steps, reducing entity conflation and improving the faithfulness of intermediate reasoning traces. Experiments across multiple multi-hop QA benchmarks demonstrate consistent gains over strong RAG and structured-RAG baselines. Looking ahead, we will study how to better construct and adapt the two-level taxonomy for specialized domains (e.g., via semi-automatic curation and domain-specific label retrieval), while keeping the hierarchy lightweight. We also plan to explore tighter integration with generation-time verification to further improve robustness under noisy or conflicting evidence.

\bibliographystyle{ACM-Reference-Format}
\bibliography{sample-base}

\clearpage
\appendix

\section{Algorithm}

We present algorithm of TaSR-RAG in Algorithm \ref{alg:TaSR-RAG}.

\begin{algorithm}[H]
  \caption{Taxonomy-Guided Structured Reasoning for Evidence Selection (TaSR-RAG)}
  \label{alg:TaSR-RAG}
  \begin{algorithmic}[1]
    \Require Query $q$, document collection $\mathcal{D}$, two-level taxonomy $\mathcal{T}$, dense-retrieval top-$K_0$, threshold $\theta$, LLMs $\text{LLM}_{\text{extract}}$, $\text{LLM}_{\text{decomp}}$, $\text{LLM}_{\text{select}}$, $\text{LLM}_{\text{answer}}$
    \Ensure Final answer $\hat{a}$ (returned as the output of the last sub-query)
    \Statex
    \State $\mathcal{D}_{\text{cand}} \gets \text{Top-}K_0\text{DenseRetrieve}(q, \mathcal{D})$
    \ForAll{$d \in \mathcal{D}_{\text{cand}}$}
    \State $\mathcal{R}_d \gets \text{LLM}_{\text{extract}}(d)$ \Comment{Document triples}
    \ForAll{$e \in \text{Entities}(\mathcal{R}_d)$}
    \State $c^{(1)} \gets \text{LLM}_{\text{select}}(\mathcal{T}^{(1)}, e)$
    \State $c^{(2)} \gets \text{LLM}_{\text{select}}(\mathcal{T}^{(2)}(c^{(1)}), e)$
    \State $\tau(e) \gets (c^{(1)}, c^{(2)})$
    \EndFor
    \State $\tilde{\mathcal{R}}_d \gets \{(\tau(h), r, \tau(t)) : (h,r,t) \in \mathcal{R}_d\}$
    \EndFor
    \Statex
    \State $\{s_1, \dots, s_N\} \gets \text{LLM}_{\text{decomp}}(q)$ \Comment{Sub-queries (triples with latent variables)}
    \ForAll{$e \in \text{Entities}(\{s_i\}_{i=1}^N)$}
    \State $c^{(1)} \gets \text{LLM}_{\text{select}}(\mathcal{T}^{(1)}, e)$
    \State $c^{(2)} \gets \text{LLM}_{\text{select}}(\mathcal{T}^{(2)}(c^{(1)}), e)$ \Comment{Query-level typing (incl. variables)}
    \State $\tau(e) \gets (c^{(1)}, c^{(2)})$
    \EndFor
    \State $\tilde{s}_i \gets (\tau(h_i), r_i, \tau(t_i))\quad \forall i$
    \Statex
    \State $\mathcal{B} \gets \emptyset$ \Comment{Entity binding table}
    \For{$i = 1$ to $N$}
    \State $s_i' \gets \text{Resolve}(s_i, \mathcal{B})$; $\tilde{s}_i' \gets \text{Resolve}(\tilde{s}_i, \mathcal{B})$
    \ForAll{$d \in \mathcal{D}_{\text{cand}}$}
    \State $S(d) \gets \text{HybridMatch}\bigl((s_i', \tilde{s}_i'), (\mathcal{R}_d, \tilde{\mathcal{R}}_d)\bigr)$
    \EndFor
    \State $\mathcal{D}_i \gets \text{Rank}(\{d \in \mathcal{D}_{\text{cand}} : S(d) \geq \theta\};\, S(d))$
    \State $\hat{o}_i \gets \text{LLM}_{\text{answer}}(s_i', \mathcal{D}_i)$
    \State $\mathcal{B} \gets \mathcal{B} \cup \text{Bind}(s_i, \hat{o}_i)$
    \EndFor
    \Statex
    \State $\hat{a} \gets \hat{o}_N$
    \State \Return $\hat{a}$
  \end{algorithmic}
\end{algorithm}

\section{Additional Experiments}

\subsection{Sensitivity to Retrieval Budget.}
\begin{figure*}[t]
  \centering
  \begin{subfigure}[b]{0.32\textwidth}
    \centering
    \includegraphics[width=\textwidth]{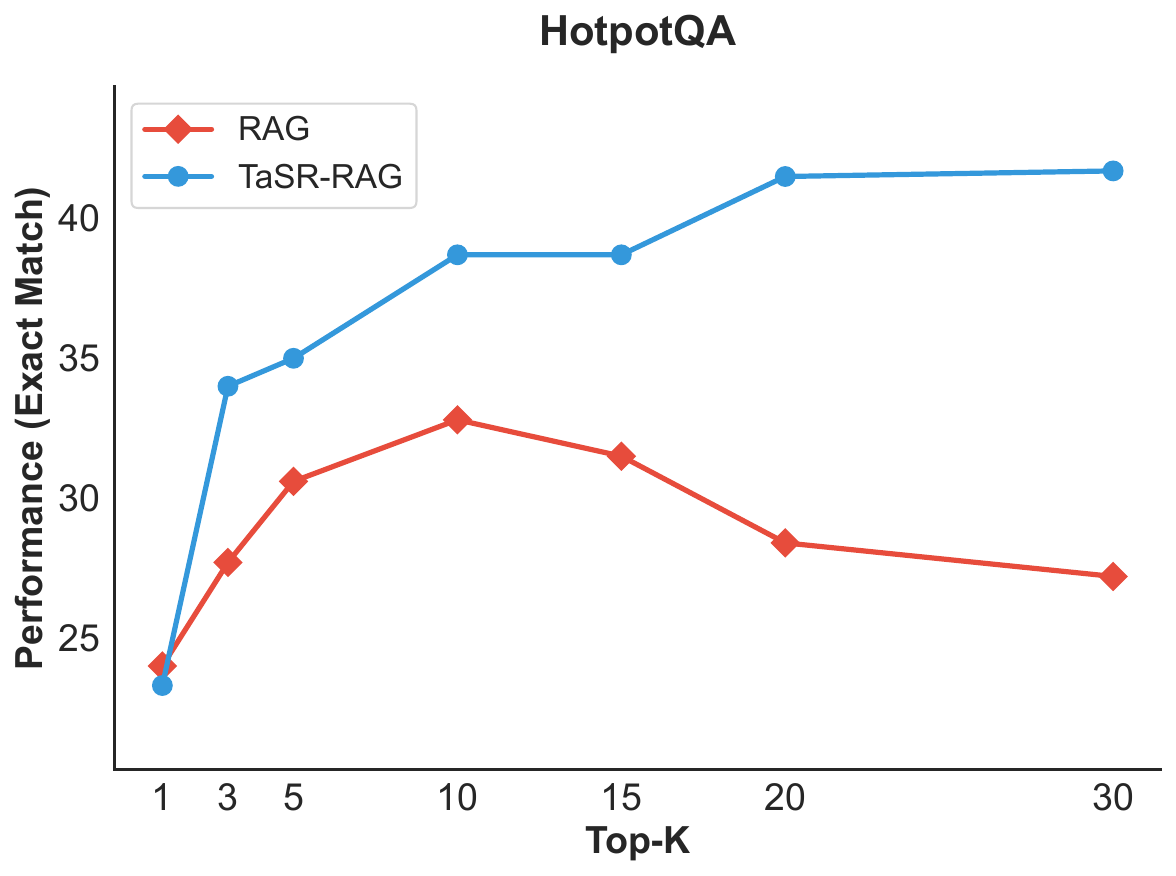}
    \caption{HotpotQA}
    \label{fig:hotpotqa}
  \end{subfigure}
  \hfill
  \begin{subfigure}[b]{0.32\textwidth}
    \centering
    \includegraphics[width=\textwidth]{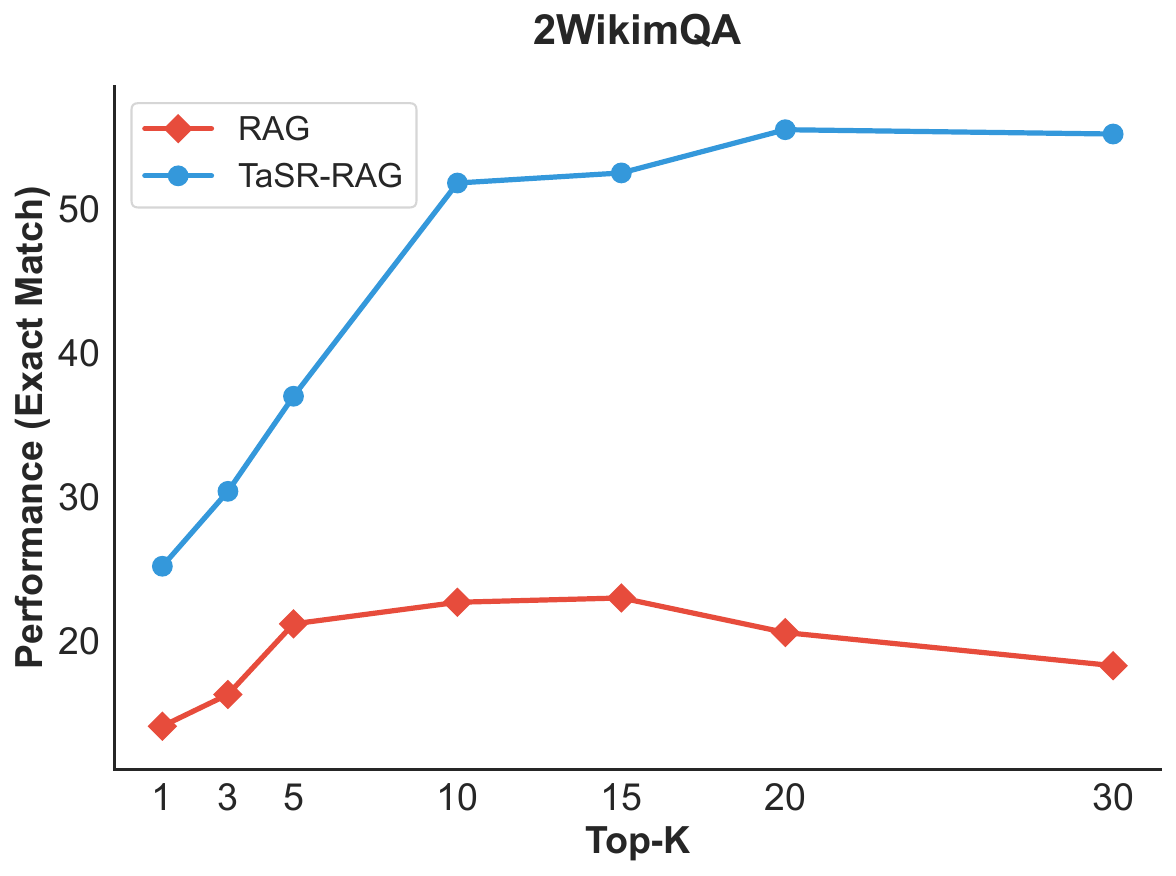}
    \caption{2WikimQA}
    \label{fig:2wikimqa}
  \end{subfigure}
  \hfill
  \begin{subfigure}[b]{0.32\textwidth}
    \centering
    \includegraphics[width=\textwidth]{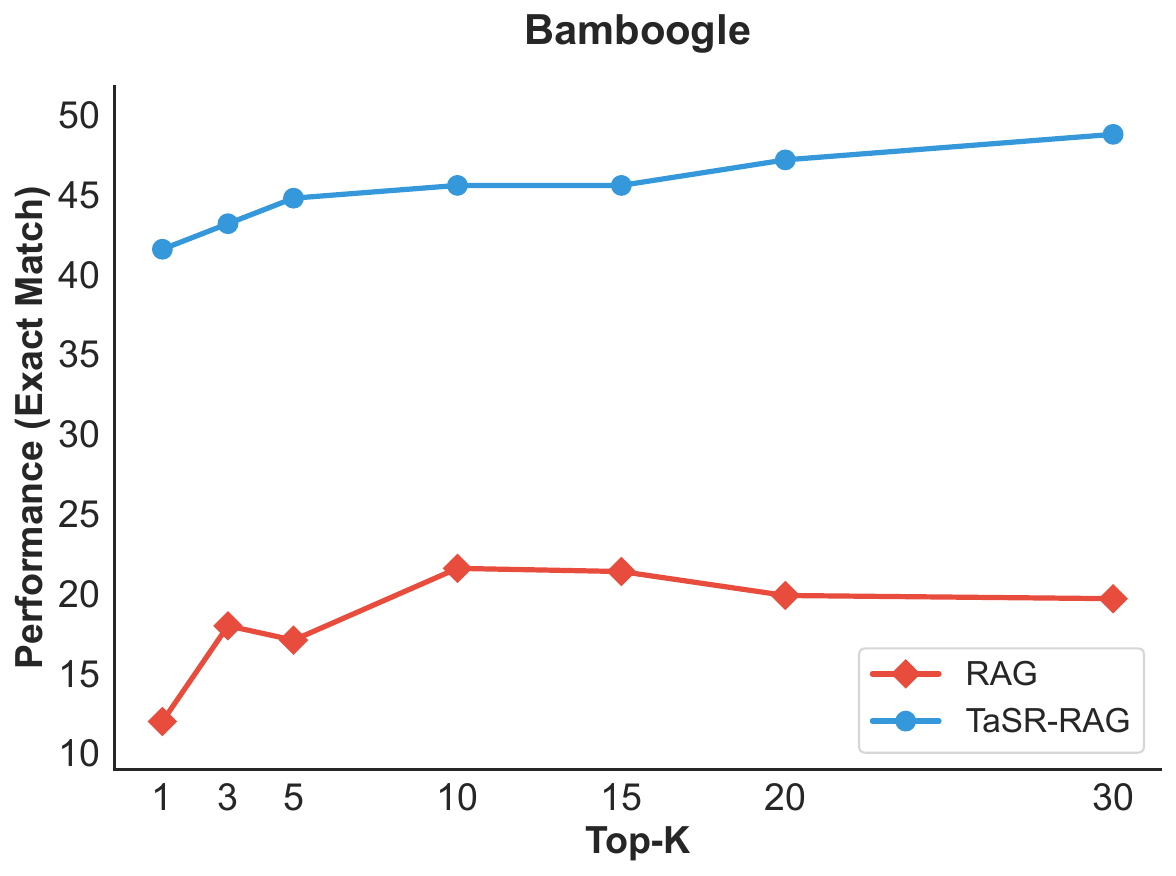}
    \caption{Bamboogle}
    \label{fig:bamboogle}
  \end{subfigure}
  \caption{Performance comparison between \textsc{TaSR-RAG} and standard RAG across different retrieval budgets (Top-K) on three multi-hop QA datasets. \textsc{TaSR-RAG} consistently outperforms RAG across all settings, with particularly strong gains on 2WikimQA (over 100\% relative improvement at Top-10) and substantial improvements on HotpotQA and Bamboogle.}
  \label{fig:topk_comparison}
\end{figure*}

We study the effect of retrieval budget by varying the number of retrieved documents as follows:
\begin{equation}
  K \in \{1, 3, 5, 10, 15, 20, 30\}.
\end{equation}
As shown in Figure~\ref{fig:topk_comparison}, \textsc{TaSR-RAG} consistently outperforms standard RAG across all retrieval budgets and datasets, demonstrating robust performance under different resource constraints.

The performance gap between \textsc{TaSR-RAG} and RAG is particularly pronounced in certain regimes.
On HotpotQA, \textsc{TaSR-RAG} achieves 38.7\% at Top-10, compared to RAG's 32.8\%, representing an 18\% relative improvement.
On 2WikimQA, the advantage is even more striking: TaSR-RAG reaches 51.8\% at Top-10 versus RAG's 22.7\%, more than doubling the baseline performance.
For Bamboogle, TaSR-RAG attains 45.6\% at Top-10 compared to RAG's 21.6\%, demonstrating substantial gains across multi-hop reasoning scenarios.

Interestingly, TaSR-RAG exhibits different scaling behaviors across datasets.
On HotpotQA and Bamboogle, performance generally improves with larger $K$, peaking around Top-15 to Top-20 before slight degradation at Top-30, suggesting that retrieval noise accumulates beyond optimal budget.
In contrast, on 2WikimQA, TaSR-RAG shows remarkable efficiency at lower budgets, with performance jumping dramatically from Top-1 (25.2\%) to Top-10 (51.8\%), then maintaining stable high performance through Top-30.
This suggests that TaSR-RAG's structured reasoning and dynamic matching mechanism enable it to identify and utilize relevant evidence more efficiently, particularly when the retrieval budget is constrained.

These results highlight TaSR-RAG's ability to extract maximal value from retrieved documents through its question decomposition and adaptive reasoning strategy, rather than simply relying on retrieving more documents.

\subsection{Performance by Reasoning Depth.}
\begin{figure*}[t]
  \centering
  \includegraphics[width=0.32\textwidth]{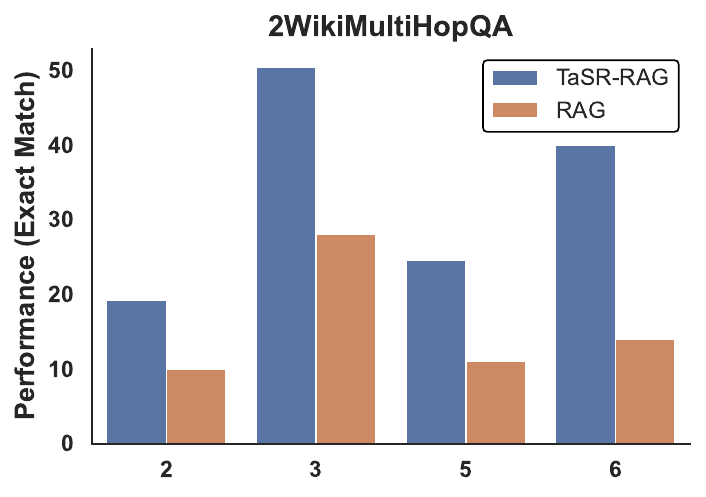}
  \hfill
  \includegraphics[width=0.32\textwidth]{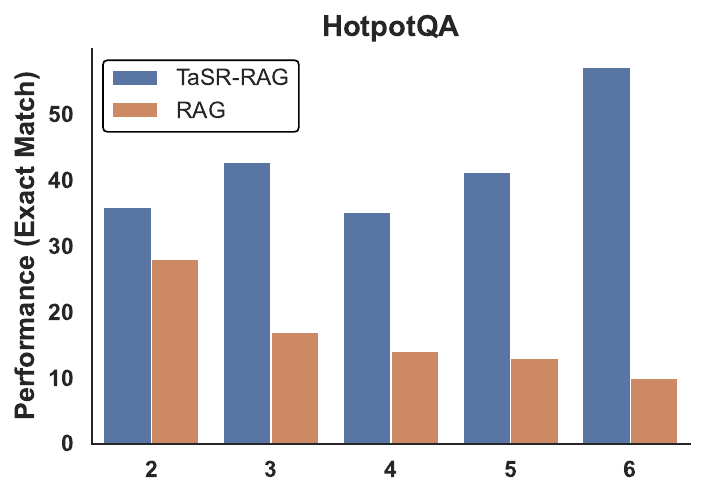}
  \hfill
  \includegraphics[width=0.32\textwidth]{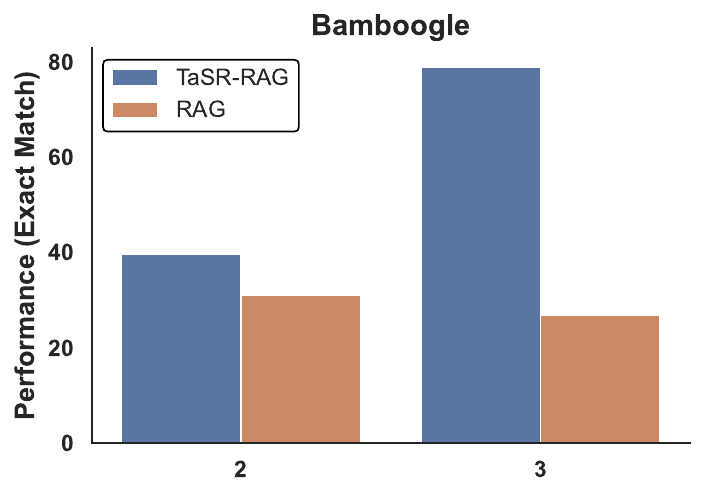}
  \caption{
    Performance by reasoning depth (hop count).
    We report results for hop counts with non-empty support.
    While RAG performance degrades as reasoning depth increases,
    \textsc{TaSR-RAG} maintains strong performance and exhibits increasingly larger gains in deeper reasoning regimes,
    with particularly pronounced improvements at 3-hop and beyond.
  }
  \label{fig:hop_ablation}
\end{figure*}

To explicitly evaluate multi-hop reasoning ability, we analyze model performance as a function of reasoning depth.
Specifically, we group test instances by hop count and report performance for hop values with non-empty support.
Figure~\ref{fig:hop_ablation} compares \textsc{TaSR-RAG} with standard RAG across three representative multi-hop QA benchmarks.

Across all datasets, both methods exhibit comparable performance on shallow reasoning cases.
However, as reasoning depth increases, standard RAG degrades rapidly, reflecting its reliance on flat retrieval and one-shot generation.
In contrast, \textsc{TaSR-RAG} maintains strong performance and shows increasingly larger gains in deeper reasoning regimes.
The advantage becomes particularly pronounced at 3-hop and beyond, where explicit sequential structure and entity binding are critical.

The trend is most evident on Bamboogle, which emphasizes implicit and indirect multi-hop inference, where \textsc{TaSR-RAG} substantially outperforms RAG at higher hop counts.
Similar patterns are observed on 2WikiMultiHopQA and HotpotQA, indicating that the benefits of taxonomy-guided sequential context selection generalize across different multi-hop reasoning styles.

\section{Two-level Entity Taxonomy}
\label{sec:appendix_taxonomy}

Table~\ref{tab:appendix_taxonomy} lists the two-level entity taxonomy used throughout \textsc{TaSR-RAG} for taxonomy-guided typing.
We use the top-level label (L1) to provide coarse semantic constraints (e.g., \texttt{PERSON}, \texttt{ORGANIZATION}), and the second-level label (L2) to capture finer distinctions (e.g., \texttt{Scientist} vs. \texttt{Politician}) that improve precision during multi-hop evidence selection.

\section{Dataset Statistics}
\label{sec:appendix_data_statistics}

Our method does not require any additional training; therefore, we only report the evaluation datasets used in our experiments. Table~\ref{tab:data_statistics} summarizes dataset sizes and task types.

\begin{table*}[t]
  \centering
  \small
  \setlength{\tabcolsep}{6pt}
  \renewcommand{\arraystretch}{1.15}
  \rowcolors{2}{gray!8}{white}
  \caption{Two-level entity taxonomy used for type constraints (L1 and L2).}
  \label{tab:appendix_taxonomy}
  \resizebox{0.98\textwidth}{!}{%
    \begin{tabularx}{\textwidth}{@{}lX@{}}
      \toprule
      \textbf{L1} & \textbf{L2 categories} \\
      \midrule
      \texttt{PERSON} & \texttt{Scientist}, \texttt{Engineer}, \texttt{Academic}, \texttt{Politician}, \texttt{Businessperson}, \texttt{Athlete}, \texttt{Actor}, \texttt{Musician}, \texttt{Writer}, \texttt{Journalist}, \texttt{Inventor}, \texttt{MilitaryPerson} \\
      \texttt{ORGANIZATION} & \texttt{Company}, \texttt{University}, \texttt{ResearchInstitute}, \texttt{GovernmentAgency}, \texttt{Nonprofit}, \texttt{InternationalOrganization}, \texttt{MilitaryUnit}, \texttt{SportsTeam}, \texttt{PoliticalParty}, \texttt{MediaOutlet}, \texttt{Hospital}, \texttt{School} \\
      \texttt{LOCATION} & \texttt{Country}, \texttt{StateOrProvince}, \texttt{City}, \texttt{Region}, \texttt{Continent}, \texttt{River}, \texttt{Lake}, \texttt{Mountain}, \texttt{Island}, \texttt{SeaOrOcean}, \texttt{Desert}, \texttt{Park} \\
      \texttt{FACILITY} & \texttt{Building}, \texttt{Bridge}, \texttt{Airport}, \texttt{Station}, \texttt{Port}, \texttt{Museum}, \texttt{Stadium}, \texttt{Campus}, \texttt{Laboratory}, \texttt{PowerPlant} \\
      \texttt{EVENT} & \texttt{War}, \texttt{Election}, \texttt{Tournament}, \texttt{Conference}, \texttt{Festival}, \texttt{Disaster}, \texttt{Protest}, \texttt{LaunchEvent}, \texttt{MergerEvent}, \texttt{Trial} \\
      \texttt{WORK} & \texttt{Book}, \texttt{Film}, \texttt{TVSeries}, \texttt{Song}, \texttt{Album}, \texttt{VideoGame}, \texttt{SoftwareProject}, \texttt{ResearchPaper}, \texttt{LawOrPolicy}, \texttt{Dataset} \\
      \texttt{PRODUCT} & \texttt{CloudService}, \texttt{Database}, \texttt{ProgrammingLanguage}, \texttt{HardwareDevice}, \texttt{VehicleModel}, \texttt{Drug}, \texttt{Chemical}, \texttt{ConsumerProduct}, \texttt{ModelOrAlgorithm} \\
      \texttt{BIOENTITY} & \texttt{Animal}, \texttt{Plant}, \texttt{Bacteria}, \texttt{Virus}, \texttt{Disease}, \texttt{ProteinOrGene} \\
      \texttt{TIME} & \texttt{Year}, \texttt{Date}, \texttt{TimePeriod} \\
      \texttt{QUANTITY} & \texttt{Count}, \texttt{Money}, \texttt{Percentage}, \texttt{Measurement} \\
      \texttt{CONCEPT} & \texttt{Technology}, \texttt{Method}, \texttt{Theory}, \texttt{FieldOfStudy}, \texttt{RoleOrTitle} \\
      \texttt{OTHER} & \texttt{Other} \\
      \bottomrule
    \end{tabularx}
  }
\end{table*}

\begin{table}[t]
  \centering
  \caption{Dataset statistics for evaluation.}
  \label{tab:data_statistics}
  \resizebox{0.98\columnwidth}{!}{%
    \begin{tabular}{llrr}
      \toprule
      \textbf{Split} & \textbf{Dataset} & \textbf{\#Examples} & \textbf{Task Type} \\
      \midrule
      \multirow{7}{*}{Eval} & NQ & 3,610 & Single-hop \\
      & TriviaQA & 11,313 & Single-hop \\
      & PopQA & 14,267 & Single-hop \\
      & HotpotQA & 7,405 & Multi-hop \\
      & 2WikiMultiHopQA & 12,576 & Multi-hop \\
      & Musique & 2,417 & Multi-hop \\
      & Bamboogle & 125 & Multi-hop \\
      \bottomrule
    \end{tabular}
  }
\end{table}

\section{Implementation Details}

We evaluate our method using Qwen2.5-7B-Instruct and Qwen2.5-72B-Instruct as the backbone generators. All components involving prompting—including query decomposition, relational triple extraction, entity typing decisions, sub-query answering, and final answer synthesis—are implemented via LLM inference. For initial dense retrieval, we retrieve top-$K_0$ candidate documents with $K_0=10$. During structured reasoning (evidence selection), the number of selected documents $K$ is dynamic and determined by score-based filtering with a threshold $\theta=0.3$. For taxonomy-guided entity typing, we use a retrieval-first, coarse-to-fine strategy: for each entity, top-$N$ first-level type candidates are retrieved using embedding similarity with $N=10$, from which the LLM selects the top-3 labels; under each selected first-level label, top-$M$ second-level candidates are retrieved with $M=20$, and the LLM selects the final $(\mathrm{L1}, \mathrm{L2})$ pair from their union. Document-level scoring aggregates over the top-$t=3$ sub-queries. Hybrid matching combines structural and semantic signals with $\alpha=0.5$; type-level matching uses $w^{(1)}=0.5$ for L1 and $w^{(2)}=0.5$ for L2, head and tail contributions use $w_h=0.5$ and $w_t=0.5$, and semantic matching weights are set to $\lambda_h=0.3$, $\lambda_r=0.3$, and $\lambda_t=0.4$. All embedding-based retrieval, including taxonomy label retrieval, uses an E5 encoder with L2-normalized embeddings and inner-product FAISS search, which is equivalent to cosine similarity. Hierarchical typing is implemented with a two-stage FAISS setup consisting of a global L1 index and separate L2 indexes for each L1 branch, enabling efficient coarse-to-fine candidate retrieval. To reduce LLM overhead, we apply simple rule-based typing for structured entities such as years, dates, and percentages before invoking embedding- or LLM-based typing. For robustness, all classification-style prompting steps use deterministic decoding (temperature $=0$), enforce JSON-only outputs, and optionally include lightweight disambiguation context constructed from document titles and evidence sentences associated with extracted triples. Unless otherwise specified, all experiments use the same settings across datasets and model scales.

\section{Case Study}
\label{sec:case_study}

We present a representative case aligned with Figure~\ref{fig:intro} to illustrate how \textsc{TaSR-RAG} performs taxonomy-guided structured reasoning for evidence selection. The question asks for two linked facts: (i) which database \textit{Science Activity Planner} uses, and (ii) which company developed that database. After dense retrieval (Step~0), the candidate pool $\mathcal{D}_{\text{cand}}$ contains heterogeneous snippets: Doc[1] directly states that \textit{Science Activity Planner} uses \textit{MySQL}, Doc[3] mentions \textit{PostgreSQL} and \textit{MySQL} only as deployment options, and Doc[6] provides a developer fact about \textit{MySQL}. \textsc{TaSR-RAG} first converts each candidate document into relational triples and taxonomy-typed triples (Step~1), then decomposes the question into an ordered chain of triple sub-queries with latent variables (Step~2).
\begin{align*}
  s_1 &= (\text{Science Activity Planner},\, \texttt{uses},\, ?x),\\
  s_2 &= (?x,\, \texttt{developed\_by},\, ?y),
\end{align*}
with corresponding typed forms that enforce the intended reasoning pattern,
\begin{align*}
  \tilde{s}_1 &= (\textsc{Work/System},\, \texttt{uses},\, \textsc{Work/Database}),\\
  \tilde{s}_2 &= (\textsc{Work/Database},\, \texttt{developed\_by},\, \textsc{Organization/Company}).
\end{align*}
This typed decomposition makes the required reasoning pattern explicit and prevents the second hop from drifting to irrelevant entities or types.

With the binding table initialized (Step~3), \textsc{TaSR-RAG} performs hop-wise evidence selection via hybrid (semantic + structural) triple matching. For hop~1 (Step~4), we compute HybridMatch scores $S(d)$ between $(s_1,\tilde{s}_1)$ and each $(\mathcal{R}_d,\tilde{\mathcal{R}}_d)$, then filter by $S(d)\ge\theta$ and rank. In this pool, Doc[1] is retained because it contains a relation-consistent and type-consistent triple supporting
\begin{equation*}
  (\text{Science Activity Planner},\, \texttt{uses},\, \text{MySQL}),
\end{equation*}
whereas Doc[3] is filtered since it does not entail a \texttt{uses} relation (it only lists options). The hop-1 answerer updates the binding table (Step~5) with $?x\leftarrow\text{MySQL}$, which turns hop~2 into the grounded query
\begin{equation*}
  s_2'=(\text{MySQL},\, \texttt{developed\_by},\, ?y).
\end{equation*}
For hop~2 (Step~6), HybridMatch is recomputed conditioned on the resolved entity, promoting Doc[6] and down-weighting any non-developer mentions of \textit{MySQL}; the final hop binds $?y\leftarrow\text{MySQL AB}$ and returns \textbf{MySQL; MySQL AB} (Step~7). This example highlights how sequential binding interacts with hybrid matching: the first hop resolves an intermediate entity with high precision, and the second hop uses that binding to retrieve developer evidence that would be hard to isolate under a single, unconstrained retrieval step.

\begin{table*}[htbp]
  \centering
  \small
  \setlength{\tabcolsep}{4pt}
  \renewcommand{\arraystretch}{1.12}
  \resizebox{\textwidth}{!}{%
    \begin{tabular}{p{1.2cm} p{5.0cm} p{7.2cm} p{2.6cm}}
      \toprule
      \multicolumn{4}{l}{\textbf{Question:} Which database does \textit{Science Activity Planner} use, and which company developed it?} \\
      \midrule
      \textbf{Step} & \textbf{Sub-query / operation} & \textbf{Retrieved/selected content} & \textbf{Binding update} \\
      \midrule

      0 & Candidate retrieval &
      Dense retrieval returns an evidence pool $\mathcal{D}_{\text{cand}}$ containing Doc[1], Doc[3], Doc[6]. & --- \\
      \midrule

      1 & Document triples + typing &
      Extract triples $\mathcal{R}_d$ and typed triples $\tilde{\mathcal{R}}_d$ for each $d\in\mathcal{D}_{\text{cand}}$. & --- \\
      \midrule

      2 & Query decomposition + typing &
      $s_1=(\text{Science Activity Planner},\,\texttt{uses},\,?x)$; $s_2=(?x,\,\texttt{developed\_by},\,?y)$.\newline
      $\tilde{s}_1=(\textsc{Work/System},\,\texttt{uses},\,\textsc{Work/Database})$; $\tilde{s}_2=(\textsc{Work/Database},\,\texttt{developed\_by},\,\textsc{Organization/Company})$. & --- \\
      \midrule

      3 & Initialize bindings &
      Initialize entity binding table $\mathcal{B}\gets\emptyset$. & $\mathcal{B}_0=\emptyset$ \\
      \midrule

      4 & Hop 1: context selection &
      Compute HybridMatch scores $S(d)$ for $d\in\mathcal{D}_{\text{cand}}$ w.r.t. $(s_1,\tilde{s}_1)$, then filter by $S(d)\ge\theta$ and rank.
      In this pool, Doc[1] is kept (relation- and type-consistent), while Doc[3] is filtered (option list; weak match).
      Select $\mathcal{D}_1=\{\text{Doc[1]}\}$. & --- \\
      \midrule

      5 & Hop 1: bind variable &
      $\hat{o}_1\gets\text{LLM}_{\text{answer}}(s_1,\mathcal{D}_1)$; bind the latent variable in $s_1$. & $\mathcal{B}_1:\ ?x\leftarrow\text{MySQL}$ \\
      \midrule

      6 & Hop 2: context selection &
      Resolve $s_2'=(\text{MySQL},\,\texttt{developed\_by},\,?y)$, then recompute $S(d)$ for all $d\in\mathcal{D}_{\text{cand}}$ conditioned on the binding.
      Filter by $S(d)\ge\theta$ and rank; Doc[6] is kept as the top match, while non-developer mentions of MySQL are down-weighted.
      Select $\mathcal{D}_2=\{\text{Doc[6]}\}$. & --- \\
      \midrule

      7 & Hop 2: bind + answer &
      $\hat{o}_2\gets\text{LLM}_{\text{answer}}(s_2',\mathcal{D}_2)$; update bindings and return $\hat{a}=\hat{o}_2$. & $\mathcal{B}_2:\ ?y\leftarrow\text{MySQL AB}$ \\
      \midrule

      \multicolumn{4}{l}{\textbf{Answer:} \textbf{MySQL; MySQL AB}} \\
      \bottomrule
    \end{tabular}
  }
  \caption{Case study aligned with Figure~\ref{fig:intro} and our method. \textsc{TaSR-RAG} performs step-wise evidence selection via hybrid (semantic + structural) triple matching and maintains explicit bindings ($\mathcal{B}_1,\mathcal{B}_2$) to carry resolved entities across hops.}
  \label{tab:case_intro_method_steps}
\end{table*}

\end{document}